\documentclass[runningheads]{llncs}

\usepackage{eccv}

\usepackage{eccvabbrv}
\usepackage{graphicx}
\usepackage{booktabs}
\usepackage{multirow}
\usepackage{amsmath}
\usepackage{amssymb}
\usepackage{enumitem}
\usepackage[accsupp]{axessibility}
\usepackage{hyperref}
\usepackage{orcidlink}

\newcommand{\compacttable}{%
    \scriptsize
    \setlength{\tabcolsep}{3pt}
    \renewcommand{\arraystretch}{0.92}
}

\usepackage[absolute]{textpos}
\definecolor{somegray}{gray}{0.6}
\newcommand{\darkgrayed}[1]{\textcolor{somegray}{#1}}

\begin{document}

\begin{textblock}{8}(4, 0.8)
\begin{center}
\darkgrayed{Accepted to the Workshop on Neuromorphic Vision (NeVi) of the European Conference on Computer Vision (ECCV), 2026.}
\end{center}
\end{textblock}

\title{Efficient Multi-Timescale Event Representations for Feed-Forward Object Detection}

\titlerunning{Efficient Multi-Timescale Event Representations\ldots}

\author{Fredrik Lundell\inst{1,2}\orcidlink{0009-0007-9857-5214} \and
Per-Erik Forss\'en\inst{1}\orcidlink{0000-0002-5698-5983} \and
M{\aa}rten Wadenb\"ack\inst{1}\orcidlink{0000-0002-0675-2794} \and Astrid Lundmark\orcidlink{0000-0001-5759-0476}\inst{2}}

\authorrunning{F. Lundell et al.}

\institute{Department of Electrical Engineering, Link\"oping University, Sweden
\email{\{firstname.lastname\}@liu.se}\and
Saab AB, Link\"oping, Sweden
}

\maketitle

\begin{abstract}
Autonomous systems require robust low-latency perception under rapidly changing scene dynamics and challenging illumination. In event cameras, object detection commonly relies on recurrent architectures to accumulate sparse temporal information over time. This work investigates how temporal information can be encoded directly within the event representation. We propose a confidence-normalized continuous multi-timescale representation based on logarithmic B-spline temporal encoding together with a geometry-aware local confidence mechanism that exploits the spatial structure of event generation. Using a fixed feed-forward EventCenterNet detector, we show that the proposed representations consistently outperform the compact CSTR representation on PEDRo and Gen1 datasets. We further introduce a polynomial approximation from recursive exponentials that enables efficient event-by-event updates while largely preserving detection performance. Our results demonstrate that carefully designed event representations can capture much of the temporal information learned through recurrent temporal modeling, providing a promising foundation for efficient feed-forward, event-driven, and future neuromorphic object detection.

\end{abstract}

\section{Introduction}
\label{sec:intro}

Reliable object detection in dynamic environments remains challenging for conventional frame-based cameras due to motion blur, limited temporal resolution, low-light conditions, and rapidly changing illumination.
Event cameras address several of these limitations by
asynchronously measuring logarithmic brightness changes with microsecond
temporal resolution~\cite{lichtsteiner2008dvs,Posch2014,Gallego_2022}. This
enables low-latency sensing, high dynamic range, and reduced motion blur, making
event cameras attractive for robotics, autonomous driving, drones, industrial
automation, and high-speed tracking.

A central challenge in detection is how to represent temporal information efficiently. Short temporal windows preserve localization accuracy but may discard weak object evidence, whereas longer windows improve temporal persistence at the cost of stale structures and reduced localization precision.

Modern detectors commonly model temporal information through recurrent hidden-state propagation or recurrent attention mechanisms~\cite{silva2024recurrentyolov8basedframeworkeventbased,Gehrig_2023_CVPR}. Although effective, these approaches introduce sequential dependencies and additional architectural complexity, limiting parallel execution and making efficient event-driven implementations more challenging.

This work investigates whether temporal information can be encoded directly within the event representation. We propose a confidence-normalized multi-timescale representation combining logarithmic B-spline encoding with geometry-aware local confidence support, preserving high resolution for recent events while compressing older temporal context. The confidence is conceptually related to the event-count support in CSTR~\cite{10254219}, but uses temporally weighted support for normalization, similar in spirit to normalized convolution~\cite{knutsson1993normalized}, with optional spatial aggregation to reinforce coherent event structures.

Exponential decay~\cite{expdecay} naturally supports recursive event-by-event updates but exhibit relatively high inter-channel correlation. In contrast, logarithmic B-spline encoding provides smoothly overlapping temporal functions with lower correlation, improving temporal diversity. To recover recursive updates, we introduce an exponential-polynomial approximation that closely matches the analytical spline representation while enabling efficient event-by-event computation.

Our main contributions are:
\begin{itemize}
\item A confidence-normalized multi-timescale event representation based on logarithmic B-spline temporal encoding.
\item An analysis of the influence of temporal encoding, temporal support, and local confidence on detection performance across the PEDRo and Gen1 datasets.
\item A polynomial approximation from recursive exponentials enabling efficient event-by-event updates of the logarithmic B-spline representation.
\item Experimental evidence that carefully designed event representations capture much of the temporal information learned through recurrent temporal modeling.
\end{itemize}

\section{Related Work}

\subsection{Event representations}
Most event-based vision methods first convert asynchronous event streams into grid-based representations that can be processed by conventional convolutional neural networks or vision transformers. Common representations include event count images, time surfaces, spatio-temporal histograms, and voxel grids.
Time surfaces encode recent event timestamps through temporal decay
functions~\cite{lagorce2017hots,sironi2018hats}, while voxel grids discretize
events into temporal bins~\cite{zhu2019evflownet}. These representations are
effective, but their temporal precision and temporal support are largely fixed
by the chosen bin width or accumulation window.

Compact image-like encodings reduce memory and computation. CSTR~\cite{10254219} combines event count, polarity, and
temporal information in a three-channel representation~\cite{10254219}. Learned
representations and recurrent matrix-based aggregation have also been explored
for asynchronous vision~\cite{gehrig2019endtoendlearningrepresentationsasynchronous,cannici2020matrixlstm,9749022}.
Representation choice can strongly affect performance and
generalization in event-based perception tasks~\cite{compareHayVisualObjectTracking}.

Spatial consistency can also reinforce coherent event structures and reduce
isolated noise. The Exponential Reduced Ordinal Surface
(EROS)~\cite{gava2022puckparallelsurfaceconvolutionkernel} uses local spatial
support to maintain stable event contours under sparse and noisy conditions.
Similarly, the proposed method incorporates confidence-weighted local support
into a continuous multi-timescale temporal representation.

Despite these advances, existing representations rely primarily on fixed temporal bins or exponential decay, leaving the influence of continuous temporal function design largely unexplored. This work addresses this question using a feed-forward detector.

\subsection{Temporal modeling for event-based detection}
Event-based object detectors often adapt frame-based architectures such as CenterNet~\cite{zhou2019objectspoints}
or YOLO~\cite{yolov8}. Since event evidence may appear intermittently, recent detectors increasingly rely on recurrent hidden states, recurrent attention, or other temporal aggregation mechanisms to accumulate information over time~\cite{silva2024recurrentyolov8basedframeworkeventbased,Gehrig_2023_CVPR,10436353}. Spiking detectors further combine event-based input with neuromorphic
computation~\cite{bodden2024spikingcenternetdistillationboostedspiking}.

These approaches demonstrate the value of temporal memory, but also increase
architectural complexity. In contrast, this work investigates how much temporal information can be encoded directly before detection, thereby isolating the design of the event input from architectural temporal modeling.

\subsection{Event-driven and neuromorphic processing}

Neuromorphic processing is particularly attractive for event cameras because computation is driven asynchronously by incoming events, enabling low-latency and energy-efficient processing of sparse event streams~\cite{ding2025neuromorphic,Posch2014}. Representations based on recursively updated temporal states naturally support this paradigm, as each incoming event can be processed with constant computational cost without recomputing a temporal window. The recursive approximation proposed in this work provides one such formulation while preserving the multi-timescale temporal structure of the analytical spline representation. This event-driven formulation is compatible with asynchronous processing and may facilitate future implementations on specialized event-driven and neuromorphic hardware platforms~\cite{davies2018loihi,furber2014spinnaker}. More generally, these observations highlight the potential of representation-level temporal modeling as an alternative to architectural temporal memory.

\section{Method}
We first review event generation and formulate multi-timescale temporal encoding using both widely used exponential decay~\cite{expdecay} and the proposed logarithmic B-spline functions. We then introduce confidence normalization with optional local confidence support, followed by an exponential-polynomial approximation that enables recursive event-by-event updates of the spline representation.
\subsection{Event Generation}
\label{sec:event_background}

An event is generated when the accumulated logarithmic brightness change
exceeds a contrast threshold,
\begin{equation}
|\Delta L(x,y,t)| \geq C,
\label{eq:event_threshold}
\end{equation}
where $\Delta L(x,y,t)$ denotes the logarithmic brightness change accumulated at pixel $(x,y)$ since its previous event, evaluated at time $t$, and $C$ is the contrast threshold.
Each event is represented as
\begin{equation}
e_i = (x_i,y_i,t_i,p_i),
\label{eq:event_definition}
\end{equation}
where $(x_i,y_i)$ is the pixel location, $t_i$ the timestamp, and
$p_i\in\{-1,+1\}$ the polarity. 

Under brightness constancy, event generation can be approximated by

\begin{equation}
\frac{dL}{dt}
=
\nabla L \cdot \mathbf{u},
\label{eq:brightness_constancy}
\end{equation}
where $\nabla L$ denotes the spatial image gradient and $\mathbf{u}$ is image-plane motion. Events are therefore generated primarily along moving object gradients.

\subsection{Confidence-Normalized Multi-Timescale Representation}
\label{seq:overview}

Given the event stream in Eq.~\eqref{eq:event_definition}, the representation
projects event age onto temporal functions. For a representation queried at time
$t$, the age
\begin{equation}
\Delta t_i = t-t_i,
\label{eq:event_age}
\end{equation}
and event $i$ contributes to temporal channel $k$ according to
\begin{equation}
w_{ik}=\phi_k(\Delta t_i),
\label{eq:basis_projection}
\end{equation}
where $\phi_k$ is the $k$-th temporal function. 

The event stream is converted into a compact $3N$-channel tensor with $N$
positive, $N$ negative, and $N$ confidence/support channels. Positive and
negative temporal evidence is accumulated as
\begin{equation}
S_k^{+}(x,y)=\sum_{i:p_i=+1}\phi_k(\Delta t_i),
\qquad
S_k^{-}(x,y)=\sum_{i:p_i=-1}\phi_k(\Delta t_i).
\end{equation}
Unsigned temporal support is
\begin{equation}
C_k(x,y)=S_k^{+}(x,y)+S_k^{-}(x,y),
\label{eq:confeq}
\end{equation}
and the polarity channels are normalized by this support:
\begin{equation}
\hat{S}_k^{+}(x,y)=
\frac{S_k^{+}(x,y)}{C_k(x,y)+\epsilon},
\qquad
\hat{S}_k^{-}(x,y)=
\frac{S_k^{-}(x,y)}{C_k(x,y)+\epsilon}.
\end{equation}

Similar to normalized convolution~\cite{knutsson1993normalized}, the
temporal support measures the amount of confidence available at each temporal
scale. Thus, the normalized polarity channels encode relative positive and
negative temporal structure, while the confidence channels in Eq.~\ref{eq:confeq} indicate how
strongly each response is supported. 

\subsection{Exponential Decay Encoding}

For exponential decay encoding~\cite{expdecay},
\begin{equation}
\phi_k(\Delta t)=
\exp\left(-\frac{\Delta t}{\tau_k}\right),
\end{equation}
where $\tau_k$ is the decay constant. Small constants emphasize recent events,
while larger constants provide longer temporal persistence. Exponential states
also support recursive event-by-event updates,
\begin{equation}
D_k(t_i)=
\exp\left(-\frac{\delta t_i}{\tau_k}\right)D_k(t_{i-1})+E_i,
\end{equation}
where $\delta t_i=t_i-t_{i-1}$ and $E_i$ is the contribution of the current
event. This is attractive for low-latency and neuromorphic processing, but
exponential functions with similar $\tau_k$ can be highly correlated across timescales.

\subsection{Log-Time B-Spline Encoding}

The proposed spline representation decomposes the event history into smoothly
overlapping logarithmic B-spline functions~\cite{unser1999splines}. Event age is mapped to
\begin{equation}
\begin{aligned}
a_i &=
\log\left(1+\frac{\Delta t_i}{\tau_0}\right), &
s_k &= \frac{k}{K-1}
\log\left(1+\frac{\tau_{\max}}{\tau_0}\right),
\end{aligned}
\end{equation}
where $\tau_0$ controls temporal resolution near recent events, while
$\tau_{\max}$ determines the temporal extent over which the spline
representation is distributed. The logarithmically distributed shifts $s_k$
define the placement of the $K$ function. The $k$-th temporal function
is then
\begin{equation}
\phi_k(\Delta t_i)=
B\!\left(a_i-s_k\right),
\label{eq:spline_function}
\end{equation}
where $B(\cdot)$ denotes the B-spline basis function. Following the standard
recursive construction~\cite{unser1999splines}, the zeroth-order B-spline is defined as
\begin{equation}
B_0(t)=
\begin{cases}
1, & t\in[0,1),\\
0, & \text{otherwise},
\end{cases}
\end{equation}
and the $n$-th order B-spline is obtained by repeated convolution ($*$).
\begin{equation}
B_n(t) = (B_0 * B_{n-1})(t)
       = \underbrace{(B_0 * B_0 * \cdots * B_0)(t)}_{n+1\ \text{times}},
\end{equation}
In this work, we use the second-order basis function $n=2$.

The logarithmic mapping allocates high temporal resolution to recent events while progressively compressing older history. Compared to exponential decay, logarithmic B-splines provide smooth interpolation with lower inter-channel correlation, yielding a more diverse temporal encoding.

\begin{figure}[t]
\centering
\setlength{\tabcolsep}{3pt}
\renewcommand{\arraystretch}{1.0}

\begin{tabular}{cc}
\includegraphics[width=0.45\linewidth,trim=0 70 0 0]{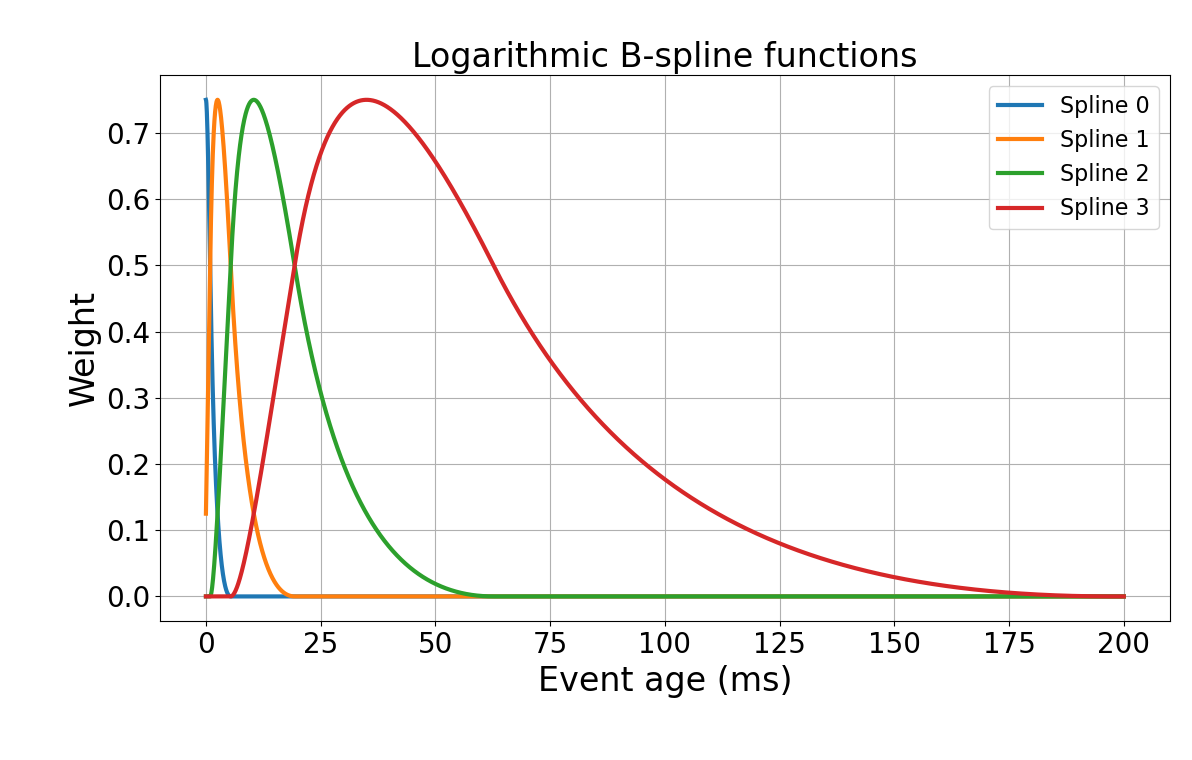} &
\raisebox{0mm}{\includegraphics[width=0.55\linewidth]{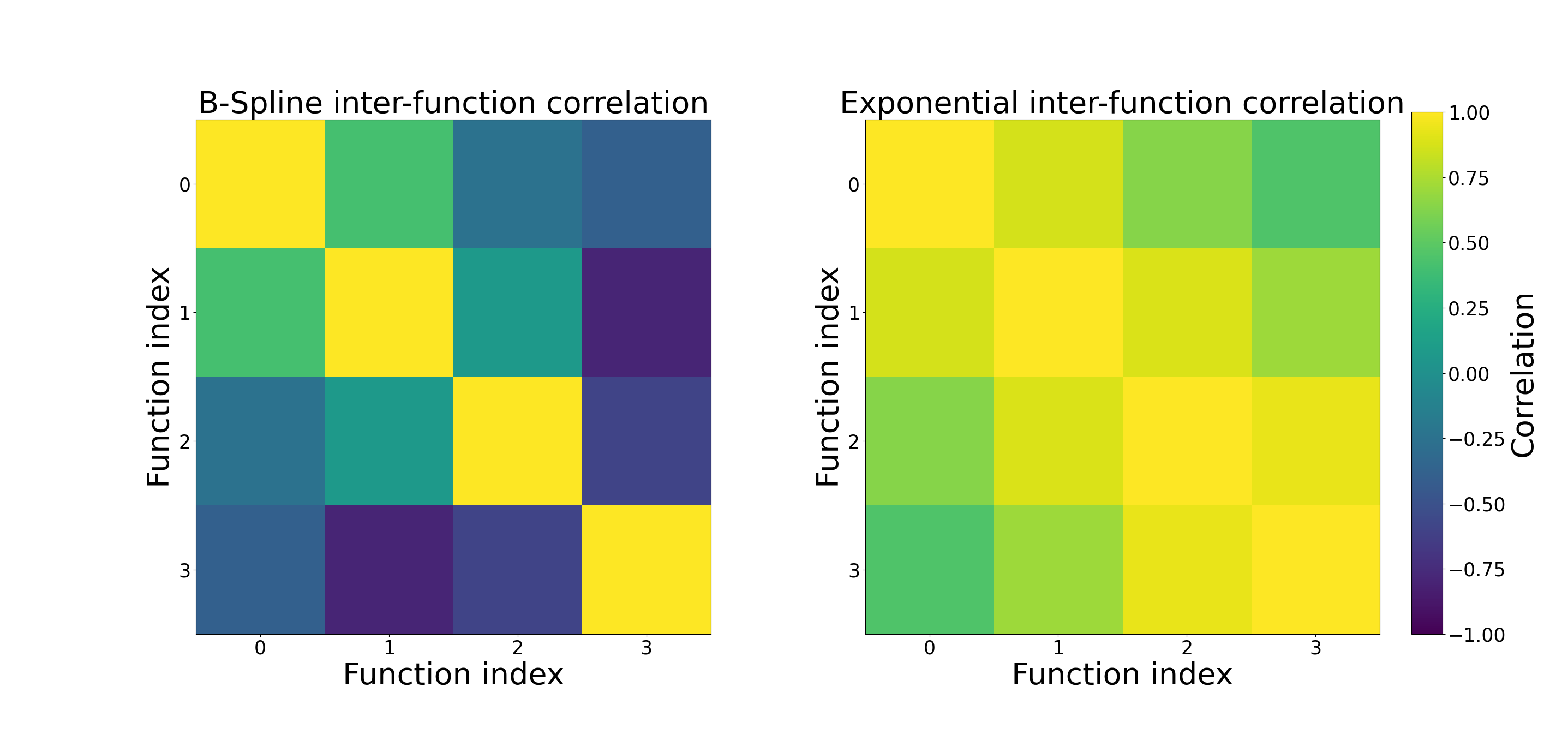}} 
\end{tabular}

\caption{
Temporal structure of the proposed logarithmic B-spline representation. \\
(a) Four Second-order logarithmic B-spline functions for $\tau_0=1.2$ and $\tau_{\max}=35$. \\
(b) Correlation matrix of logarithmic B-spline (left) and exponential decay (right) temporal
functions, showing that the B-spline functions produce a less redundant
multi-timescale encoding.
}
\label{fig:spline_basis_and_corr}
\end{figure}

Figure~\ref{fig:spline_basis_and_corr} illustrates the logarithmic B-spline
functions and compares their temporal correlations with exponential decay.

\subsection{Spatial-Neighbourhood Confidence}

Since events are generated primarily along moving structures, neighbouring
temporally correlated events provide geometric support for weak or fragmented
object evidence. Given $C_k(x,y)$, local confidence is accumulated as
\begin{equation}
\tilde{C}_k(x,y)=
\sum_{(u,v)\in\mathcal{N}(x,y)}
G(u-x,v-y, \sigma)C_k(u,v),
\end{equation}
where $\mathcal{N}(x,y)$ is a local spatial neighbourhood and $G$ is a normalized Gaussian weighting kernel, with std $\sigma$.

The confidence channels measure local spatio-temporal support rather than
polarity. This differs from smoothing the positive and negative channels:
local confidence reinforces the support used for normalization without blurring
polarity-specific temporal information.

\subsection{Polynomial Approximation of Log-Time B-Spline functions}
\label{seq:approx}

Direct spline accumulation does not naturally support constant-cost recursive
updates. To enable event-driven updates, spline functions are
approximated using polynomial combinations of exponential kernels,
\begin{equation}
e_j(\Delta t)=
\exp\left(-\frac{\Delta t}{\tau_j}\right).
\end{equation}
Each spline function in Eq.~\ref{eq:spline_function} is approximated as
\begin{equation}
\hat{B}_k(\Delta t)=
f_k(e_1(\Delta t),e_2(\Delta t),\ldots,e_M(\Delta t)),
\end{equation}
where $f_k$ is a learned polynomial mapping. Coefficients are obtained by
least-squares regression over sampled temporal points.

\begin{figure}[htbp]
\centering
\includegraphics[width=0.95\linewidth,trim=0 50pt 0 0]{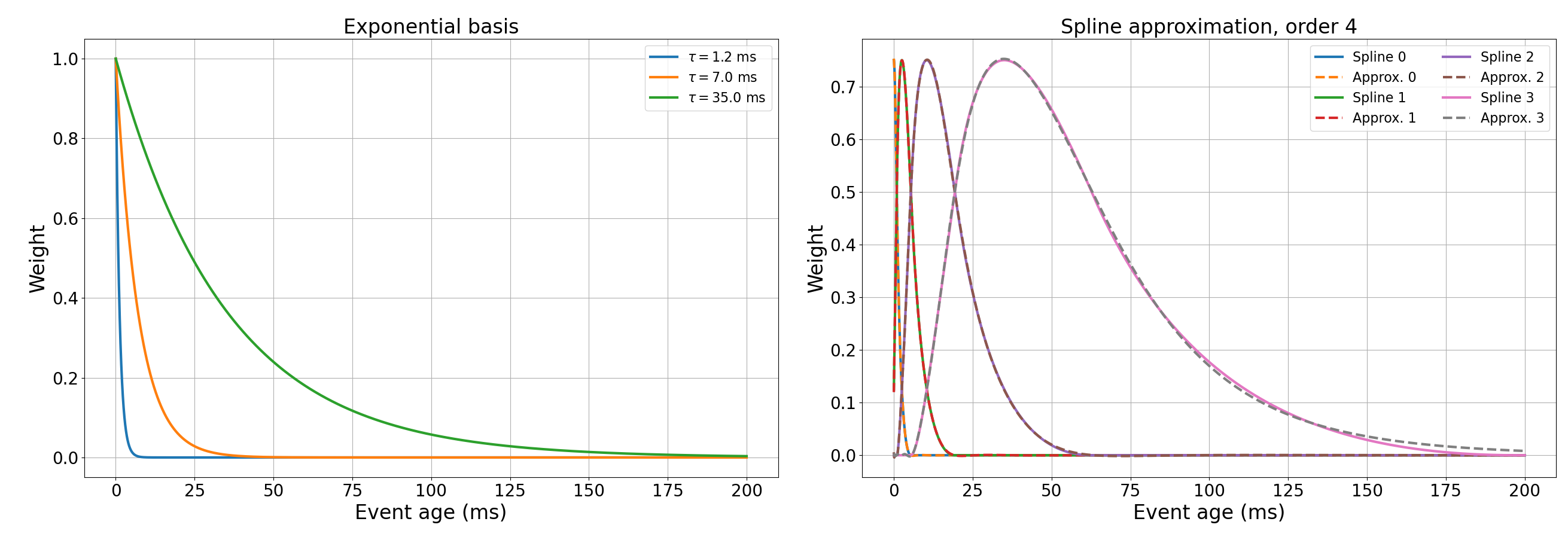}
\caption{Approximated logarithmic B-spline functions. (a) Input exponential decay kernel functions (b) Learned 4-order polynomial mapping without a bias term.}
\label{fig:splin_app_basis}
\end{figure}

Figure~\ref{fig:splin_app_basis} illustrates the fourth-order polynomial
approximation of the logarithmic B-spline functions using three exponential
functions and a learned polynomial mapping.

The exponential states can be updated recursively as events arrive, while the
polynomial mapping reconstructs spline-like responses when applied. 
It has not escaped our notice that the change of basis can be derived analytically using Gram-matrices formed by dot-products of the involved basis functions. We leave this, with efficient recursive implementation and evaluation, for future work.

\section{Experimental Setup}

Experiments are conducted on PEDRo~\cite{bbpprsPedro2023} and
Gen1~\cite{de2020large}. PEDRo consists primarily of pedestrian detection
sequences recorded from a moving platform and is dominated by medium and large
objects, with few small-object instances. Gen1 contains urban driving
sequences with a much larger proportion of small and distant objects, resulting
in sparser event observations and a more challenging detection setting. Therefore, the PEDRo and Gen1 datasets complement each other. 

\subsection{Detector and Evaluation}
\label{sec:detector}
All experiments use the proposed fully feed-forward EventCenterNet detector. The architecture consists of a ResNet34 encoder, a U-Net-style decoder with attention skip connections, and CenterNet prediction heads~\cite{resnet,UNET,attUNET,zhou2019objectspoints}. It predicts object center heatmaps, bounding-box sizes, and center offsets. No recurrent hidden state or temporal attention is used, allowing the effect of representation design to be evaluated independently. To accommodate different event representations, only the first convolutional layer is modified to match the number of input channels; the rest of the architecture is identical. A custom CenterNet with ResNet34 was chosen for architectural flexibility and fewer licensing constraints, while providing a moderate-capacity feed-forward baseline for isolating the impact of event representation design. 

Within each dataset, all models use identical training settings; only the event
representation is changed. Standard event-based augmentation is applied,
including horizontal flips, polarity flips, and timestamp jittering.

Performance is evaluated using the standard Common Objects in Context (COCO) metrics~\cite{coco}, AP$_{50:95}$ denotes Average Precision (AP) averaged over IoU thresholds from 0.50 to 0.95. We additionally report AP$_{50}$, AP$_{75}$, AP$_s$, AP$_m$, AP$_l$, and
AR$_s$, where AR denotes Average Recall and the subscripts $s$, $m$, and $l$
refer to small, medium, and large objects, respectively.

\section{Results}

\subsection{Representation Comparison}
\label{seq:Representation}

\begin{table}[t]
\centering
\caption{
Comparison of temporal representations on PEDRo and Gen1. PEDRo models use a
40~ms event window, while Gen1 models use a 50~ms window. Only the event
representation is changed within each dataset. Four temporal functions are used for both
exponential and spline-based representations.}
\compacttable
\begin{tabular}{llcccccccc}
\toprule
Dataset & Representation &
Ch. &
AP$_{50:95}$ &
AP$_{50}$ &
AP$_{75}$ &
AP$_s$ &
AP$_m$ &
AP$_l$ &
AR$_s$ \\
\midrule

\multirow{5}{*}{PEDRo}
& CSTR~\cite{10254219} & 3
& 0.566 & 0.877 & 0.648 & 0.010 & 0.563 & 0.597 & 0.072 \\
& Exp. decay & 12
& 0.621 & 0.907 & 0.739 & 0.015 & 0.593 & \textbf{0.674} & 0.142 \\
& Exp. decay + local conf. & 12
& 0.621 & 0.908 & 0.734 & 0.018 & 0.600 & 0.668 & 0.106 \\
& Spline & 12
& 0.625 & 0.907 & 0.738 & 0.041 & 0.606 & 0.671 & 0.151 \\
& Spline + local conf. & 12
& \textbf{0.630} & \textbf{0.915} & \textbf{0.752}
& \textbf{0.062} & \textbf{0.611} & 0.670 & \textbf{0.219} \\

\midrule

\multirow{5}{*}{Gen1}
& CSTR~\cite{10254219} & 3
& 0.355 & 0.637 & 0.349 & 0.278 & 0.435 & 0.274 & 0.446 \\
& Exp. decay & 12
& 0.365 & 0.646 & 0.360 & 0.285 & 0.453 & 0.307 & \textbf{0.505} \\
& Exp. decay + local conf. & 12
& 0.365 & 0.642 & 0.361 & 0.282 & 0.451 & 0.304 & 0.499 \\
& Spline & 12
& \textbf{0.370} & \textbf{0.658} & \textbf{0.361}
& \textbf{0.285} & \textbf{0.455} & 0.298 & 0.501 \\
& Spline + local conf. & 12
& 0.365 & 0.646 & 0.352 & 0.282 & 0.451 & \textbf{0.318} & 0.498 \\
\bottomrule
\end{tabular}
\label{tab:representation_comparison}
\end{table}

We compare the CSTR~\cite{10254219}, multi-timescale exponential decay and logarithmic B-spline
representations using the feed-forward EventCenterNet detector. 

Table~\ref{tab:representation_comparison} summarizes the results. All models
use four temporal functions. Exponential decay employs
$\tau=\{1.2,3.5,10,35\}$~ms, while the spline-based representation uses
four second-order functions spanning the same temporal range. PEDRo uses a
40~ms event window and Gen1 a 50~ms event window.

A normalized 3×3 binomial kernel, approximating a Gaussian filter, was used for local confidence.

Both multi-timescale representations outperform the compact CSTR baseline on
PEDRo and Gen1 across all evaluated metrics. On PEDRo, spline encoding with local
confidence achieves the highest AP$_{50:95}$, AP$_{50}$, AP$_{75}$, AP$_s$,
AP$_m$, and AR$_s$. Compared with exponential decay, AP$_s$ increases from
0.015 to 0.062 and AR$_s$ from 0.142 to 0.219. On Gen1, spline encoding without
local confidence achieves the highest AP$_{50:95}$, AP$_{50}$, and
AP$_{75}$, while local confidence primarily improves AP$_l$.
The effect of local confidence also differs between the two temporal
representations. For exponential decay, local confidence provides little or no
benefit on either dataset, whereas it consistently improves spline encoding on
PEDRo. 

\begin{figure}[t]
    \centering
    \setlength{\tabcolsep}{2pt}
    \renewcommand{\arraystretch}{1.0}

    \begin{tabular}{cccccc}
        \multicolumn{2}{c}{\textbf{CSTR}} &
        \multicolumn{2}{c}{\textbf{Exp}} &
        \multicolumn{2}{c}{\textbf{Spline + local conf.}} \\

        \textbf{Repr.} & \textbf{Heatmap} &
        \textbf{Repr.} & \textbf{Heatmap} &
        \textbf{Repr.} & \textbf{Heatmap} \\

        \includegraphics[width=0.155\textwidth]{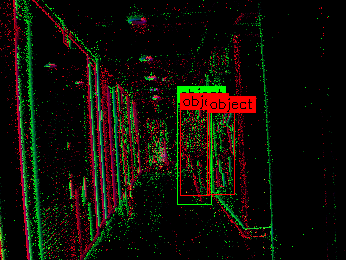} &
        \includegraphics[width=0.155\textwidth]{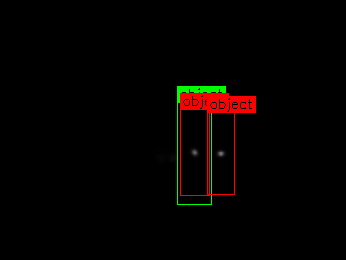} &
        \includegraphics[width=0.155\textwidth]{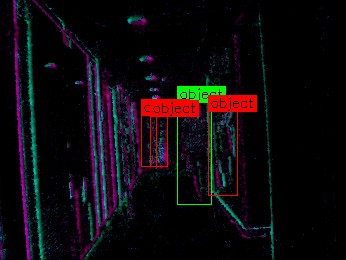} &
        \includegraphics[width=0.155\textwidth]{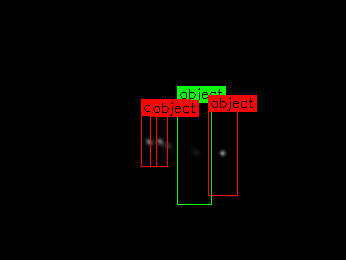} &
        \includegraphics[width=0.155\textwidth]{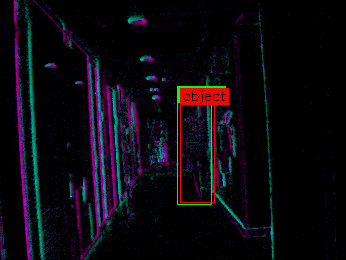} &
        \includegraphics[width=0.155\textwidth]{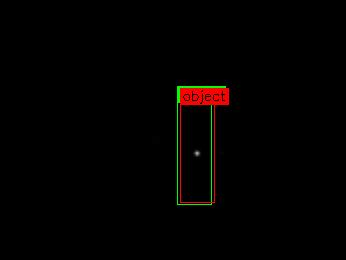} \\

        \includegraphics[width=0.155\textwidth]{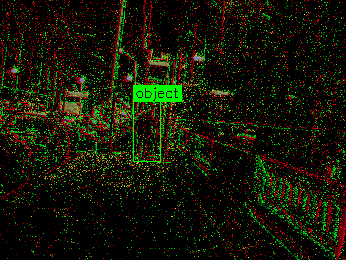} &
        \includegraphics[width=0.155\textwidth]{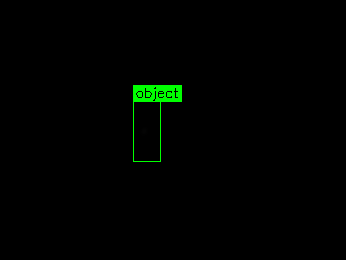} &
        \includegraphics[width=0.155\textwidth]{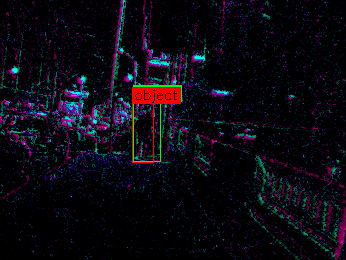} &
        \includegraphics[width=0.155\textwidth]{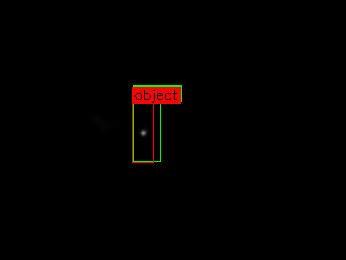} &
        \includegraphics[width=0.155\textwidth]{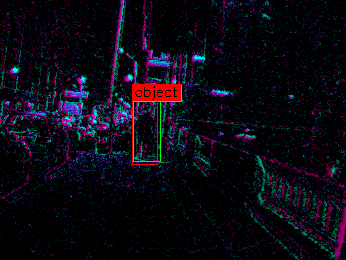} &
        \includegraphics[width=0.155\textwidth]{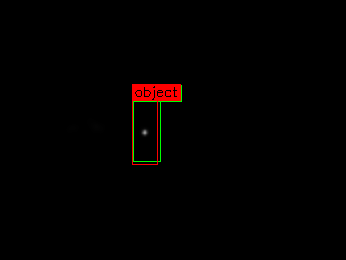} \\

        \includegraphics[width=0.155\textwidth]{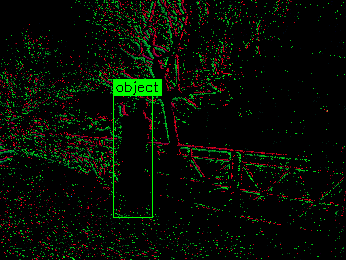} &
        \includegraphics[width=0.155\textwidth]{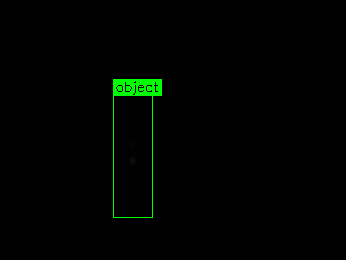} &
        \includegraphics[width=0.155\textwidth]{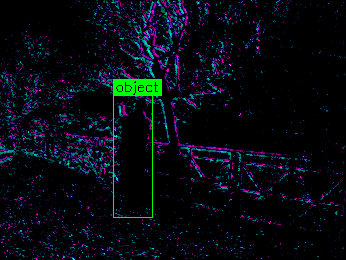} &
        \includegraphics[width=0.155\textwidth]{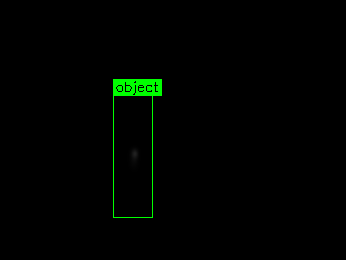} &
        \includegraphics[width=0.155\textwidth]{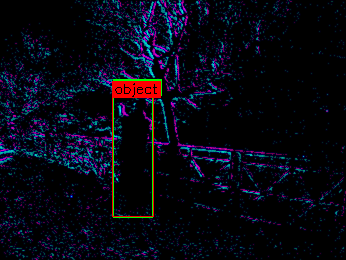} &
        \includegraphics[width=0.155\textwidth]{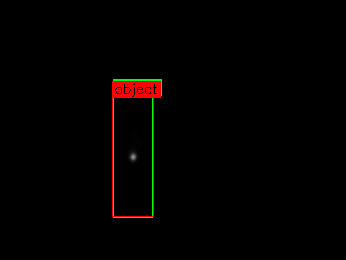}
    \end{tabular}

    \caption{Qualitative comparison of CSTR, exponential decay, and logarithmic spline representations on PEDRo. Each row shows one test example with predictions (red), annotations (green), and detector heatmaps. Exponential and spline representations are averaged over four temporal scales; red/green encode polarity support and blue local temporal confidence.}
    \label{fig:pedro_cstr_exp_spline_comparison}
\end{figure}

Figure~\ref{fig:pedro_cstr_exp_spline_comparison} presents qualitative examples from PEDRo, where the proposed spline-based representation  detects object instances that are either missed or less accurately localized by the CSTR and exponential decay representations.

\subsection{Temporal Function Placement}
\label{seq:TemporalPlacement}

To better understand the influence of temporal representation on Gen1, we investigate the effect of the window size and temporal function placement. The temporal resolution near recent events controlled by $\tau_0$, and the maximum temporal range determined by $\tau_{\max}$.

\begin{table}[t]
\centering
\caption{
Influence of window size and temporal function placement on Gen1.
All models use four second-order logarithmic B-spline functions with local confidence support. All models were trained for 20 epochs using the same training configuration.
}
\compacttable
\begin{tabular}{ccccccc}
\toprule
Window size (ms) &
$\tau_0$ (ms) &
$\tau_{\max}$ (ms) &
AP$_{50:95}$ &
AP$_{50}$ &
AP$_{75}$ &
AP$_s$ \\
\midrule
50  & 1.2 & 35 & 0.365 & 0.646 & 0.352 & 0.282 \\
150 & 1.2 & 75 & 0.387 & 0.678 & 0.386 & 0.318 \\
300 & 1.2 & 90 & 0.389 & 0.673 & 0.392 & 0.324 \\

\midrule

150 & 1.2  & 75 & 0.387 & 0.678 & 0.386 & 0.318 \\
150 & \textbf{0.6} & \textbf{75}
& \textbf{0.406}
& \textbf{0.691}
& \textbf{0.409}
& \textbf{0.335} \\
150 & 0.45 & 75 & 0.399 & 0.680 & 0.402 & 0.321 \\
150 & 0.6  & 50 & 0.400 & 0.690 & 0.401 & 0.332 \\
150 & 0.45 & 50 & 0.401 & 0.691 & 0.398 & 0.320 \\

\bottomrule
\end{tabular}
\label{tab:temporalparametergen1}
\end{table}
Comparisons in Table~\ref{tab:temporalparametergen1} show that increasing the
window size from 50~ms to 150~ms and temporal range $\tau_{\max}$ yields a substantial improvement,
demonstrating the importance of longer temporal context for sparse event
observations. Extending the window size to 300~ms with $\tau_{\max}$ = 90, provides only marginal additional
benefit, suggesting that a window size of 150~ms and $\tau_{\max}$ = 75 is sufficient for the evaluated
representation.

Temporal function placement also has a strong influence on performance. Keeping
the window size fixed at 150~ms, reducing the temporal resolution parameter
$\tau_0$ from 1.2~ms to 0.6~ms improves AP across all reported
metrics, indicating that Gen1 benefits from higher temporal resolution near
recent events. Further reducing $\tau_0$ to 0.45~ms provides no additional
improvement. Finally, reducing the temporal range $\tau_{\max}$ from 75~ms to 50~ms produces only minor performance changes. 

\begin{figure}[t]
    \centering
    \setlength{\tabcolsep}{2pt}
    \renewcommand{\arraystretch}{1.0}

    \begin{tabular}{cccccc}
        \textbf{Mean repr.} &
        \textbf{Heatmap} &
        $\boldsymbol{\tau_0}$ &
        $\boldsymbol{\tau_1}$ &
        $\boldsymbol{\tau_2}$ &
        $\boldsymbol{\tau_3}$ \\

        \includegraphics[width=0.155\textwidth]{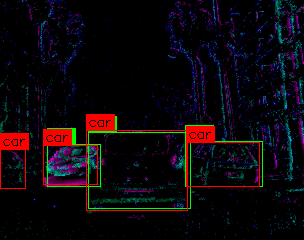} &
        \includegraphics[width=0.155\textwidth]{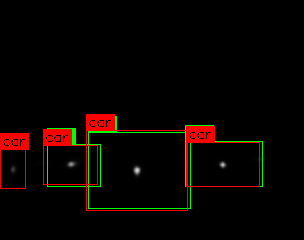} &
        \includegraphics[width=0.155\textwidth]{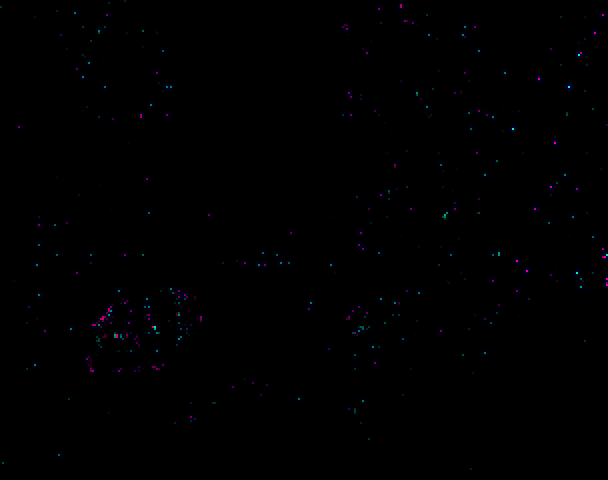} &
        \includegraphics[width=0.155\textwidth]{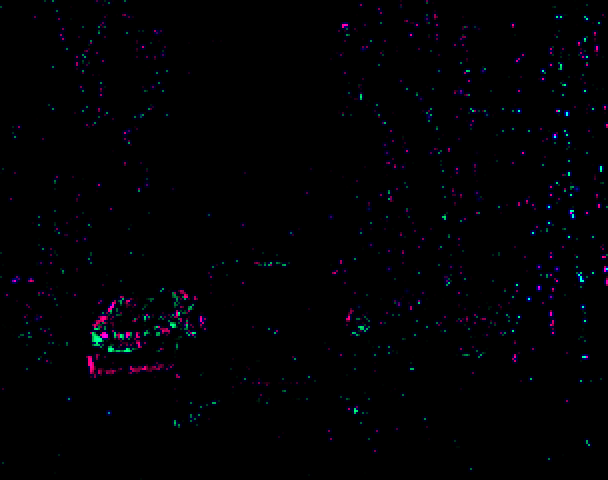} &
        \includegraphics[width=0.155\textwidth]{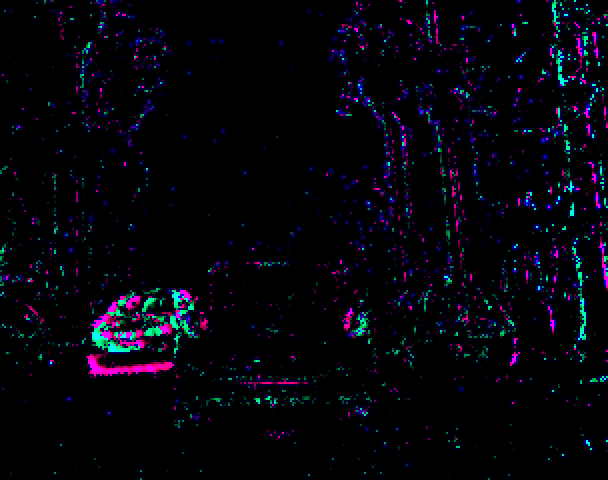} &
        \includegraphics[width=0.155\textwidth]{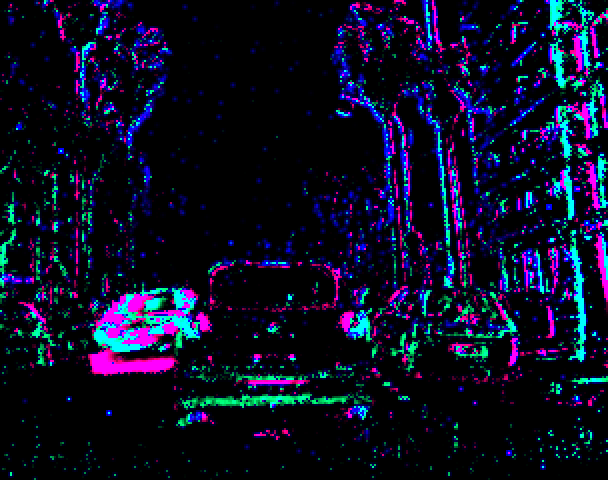} \\

        \includegraphics[width=0.155\textwidth]{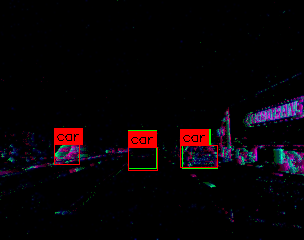} &
        \includegraphics[width=0.155\textwidth]{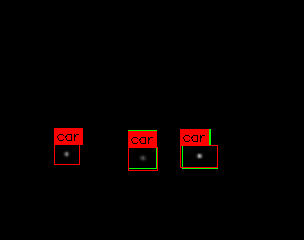} &
        \includegraphics[width=0.155\textwidth]{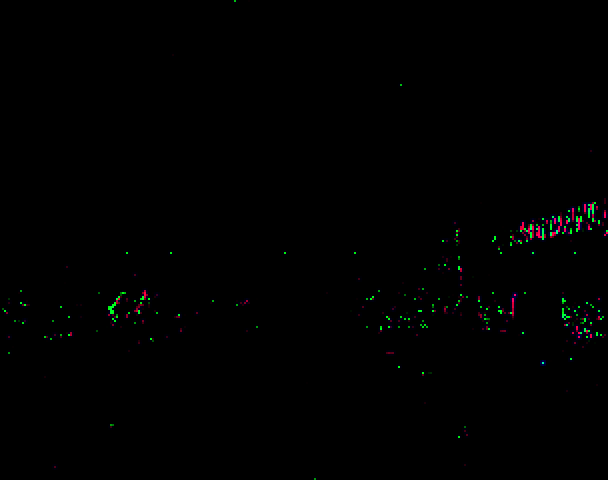} &
        \includegraphics[width=0.155\textwidth]{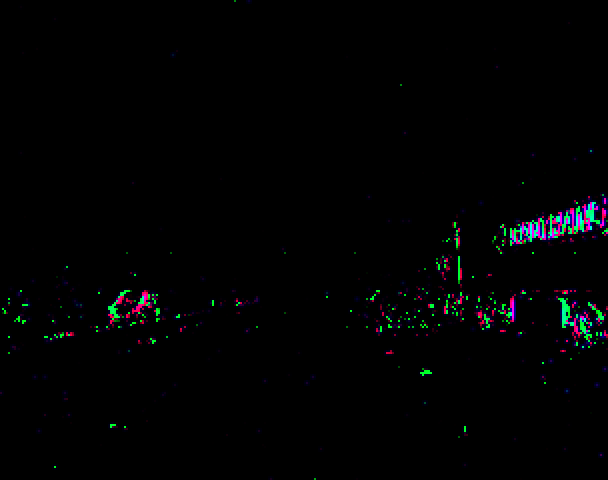} &
        \includegraphics[width=0.155\textwidth]{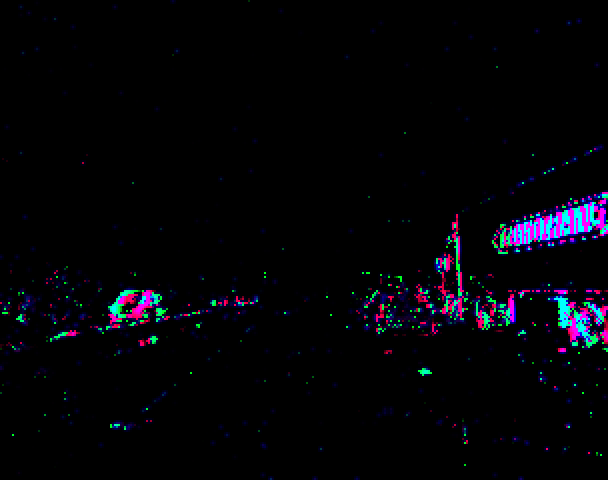} &
        \includegraphics[width=0.155\textwidth]{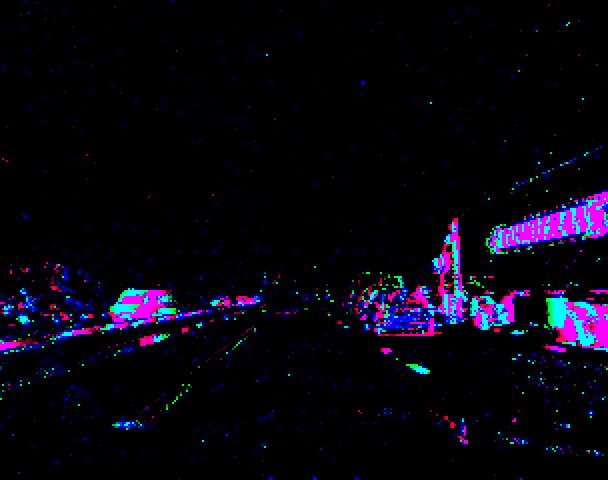} \\

        \includegraphics[width=0.155\textwidth]{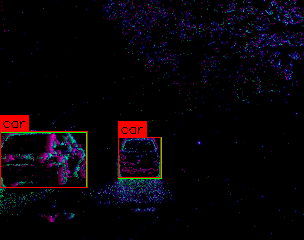} &
        \includegraphics[width=0.155\textwidth]{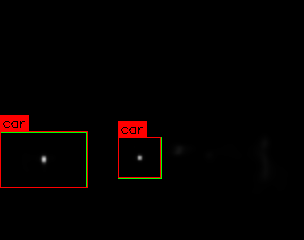} &
        \includegraphics[width=0.155\textwidth]{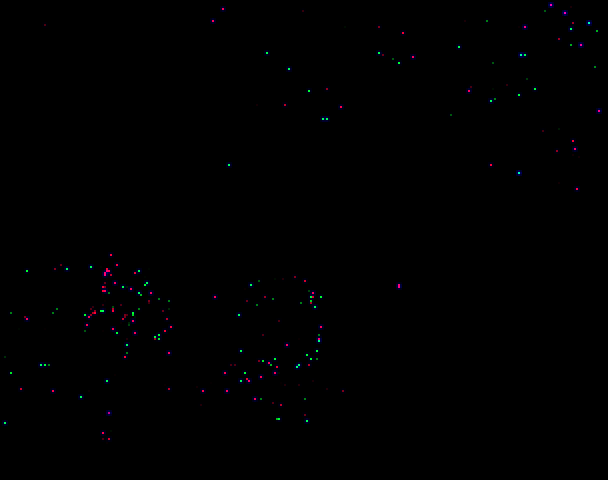} &
        \includegraphics[width=0.155\textwidth]{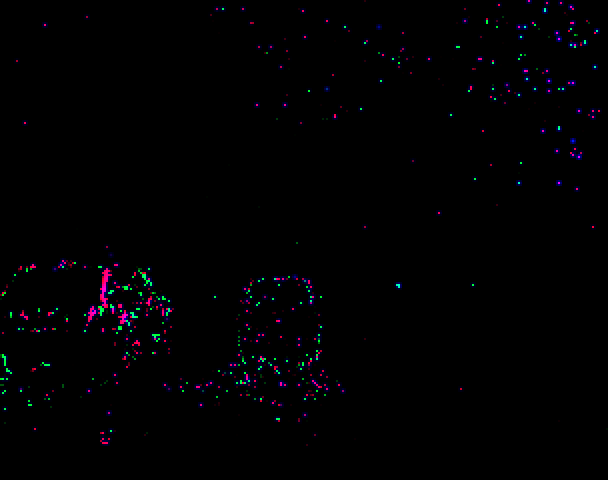} &
        \includegraphics[width=0.155\textwidth]{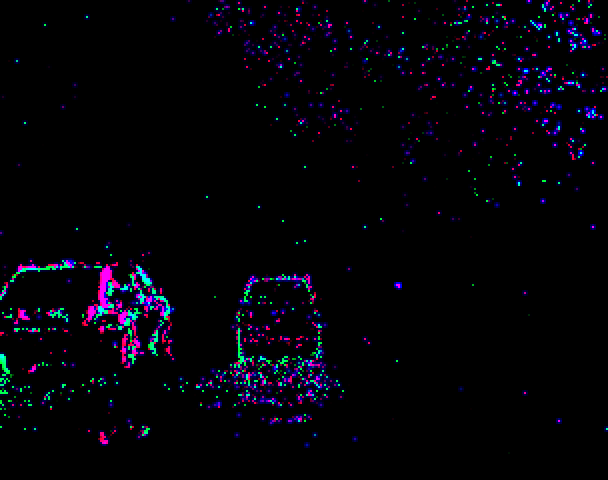} &
        \includegraphics[width=0.155\textwidth]{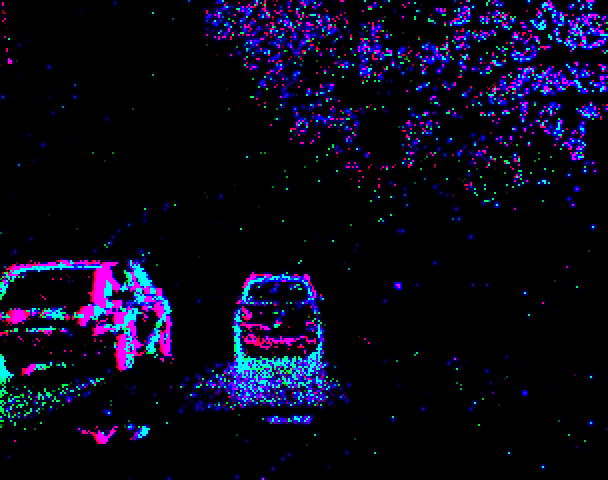}
    \end{tabular}

    \caption{Qualitative visualization of the logarithmic spline representation on Gen1 using a 150~ms event window. Columns show the averaged representation with predictions (red) and annotations (green), the detector heatmap, and the four spline channels spanning 0.6--75~ms. Red/green denote confidence-weighted polarity support and blue denotes local temporal confidence.}
    \label{fig:gen1_spline_channels}
\end{figure}

Figure~\ref{fig:gen1_spline_channels} presents qualitative Gen1 examples spanning fast highway and slower urban driving. The four temporal functions capture complementary motion dynamics across multiple timescales. Several apparent false positives, including approaching vehicles at high-speed and a parked car, are, in fact, correct detections missing from the annotations.

\begin{table}[t]
\centering
\caption{
Influence of temporal function placement on PEDRo. All models use four
second-order logarithmic B-spline functions and local confidence support.
All models were trained for 120 epochs using the same training configuration.
}
\compacttable
\begin{tabular}{ccccccc}
\toprule
Window size (ms) &
$\tau_0$ (ms) &
$\tau_{\max}$ (ms) &
AP$_{50:95}$ &
AP$_{50}$ &
AP$_{75}$ &
AP$_s$ \\
40 & 0.6 & 20 & \textbf{0.633} & 0.913 & 0.747 & 0.024 \\
40 & 0.6 & 35 & 0.627 & 0.906 & 0.750 & 0.002 \\

40 & 1.2 & 20 & 0.620 & 0.904 & 0.733 & 0.021 \\
40 & 1.2 & 35 & 0.630 & \textbf{0.915} & \textbf{0.752} & \textbf{0.062} \\
40 & 2.4 & 20 & 0.631 & 0.911 & 0.745 & 0.032 \\
40 & 2.4 & 35 & 0.628 & 0.910 & 0.747 & 0.025 \\
\bottomrule
\end{tabular}
\label{tab:pedro_temporalparameter}
\end{table}

Table~\ref{tab:pedro_temporalparameter} presents a similar parameter study on
PEDRo using a fixed 40~ms window size. While $\tau_0=0.6$~ms and
$\tau_{\max}=20$~ms achieve the highest AP$_{50:95}$, the best AP$_{50}$,
AP$_{75}$, and AP$_s$ are obtained using $\tau_0=1.2$~ms and
$\tau_{\max}=35$~ms. Small-object performance is particularly sensitive to the
shortest temporal scale, decreasing when $\tau_0$ is either reduced to
0.6~ms or increased to 2.4~ms. 

\subsection{Polynomial Approximation}
\begin{table}[t]
\centering
\caption{
Performance of recursive exponential-polynomial approximation of the
logarithmic B-spline representation on PEDRo.
}
\compacttable
\resizebox{\linewidth}{!}{%
\begin{tabular}{lcccccccc}
\toprule
Method &
Coeff. &
MSE $\downarrow$ &
AP$_{50:95}$ &
AP$_{50}$ &
AP$_{75}$ &
AP$_s$ &
AP$_m$ &
AP$_l$ \\
\midrule
Analytical spline
& Exact & --
& \textbf{0.636} & \textbf{0.918} & \textbf{0.752}
& \textbf{0.111} & \textbf{0.618} & \textbf{0.674} \\

2nd order approx.
& 9 & $1.44{\times}10^{-5}$
& 0.624 & 0.914 & 0.739 & 0.091 & 0.603 & 0.665 \\

4th order approx.
& 34 & $\mathbf{1.03{\times}10^{-6}}$
& 0.626 & 0.913 & 0.739 & 0.087 & 0.607 & 0.667 \\
\bottomrule
\end{tabular}%
}
\label{tab:approx}
\end{table}

Table~\ref{tab:approx} evaluates polynomial approximation from recursive exponentials
of the analytical spline representation. The detector is trained
once using the analytical spline representation for 200 epochs, after which the
analytical spline functions are replaced at inference by 
approximations using 9 and 34 polynomial coefficients per spline. To enable
fully recursive event-by-event updates, the polynomial mapping is fitted
without a constant bias term, allowing all representation channels to be
reconstructed solely from recursively updated exponential states.
Increasing the approximation order from 2 to 4 reduces the approximation MSE
from $1.44\times10^{-5}$ to $1.03\times10^{-6}$, while detection performance
changes only marginally. These results indicate that accurate object detection
does not require an exact reconstruction of the analytical spline function and
that a compact approximation preserves most of the discriminative
temporal information.

Additional fitting experiments on Gen1 further show that the exponential basis
should be matched to the temporal support of the analytical spline
representation. Using three exponential functions yields a total approximation
MSE of $1.89\times10^{-2}$, whereas five functions with decay constants
$\tau=\{0.6,2.5,10,40,120\}$~ms reduce the MSE to
$1.36\times10^{-6}$. This demonstrates that selecting exponential functions consistent with the temporal range of the splines is important for an accurate approximation.

\subsection{Comparison with Feed-Forward and Recurrent Detectors}

\begin{table}[t]
\centering
\caption{
Comparison with feed-forward and recurrent YOLO-based detectors. ReYOLO AP$_{50}$ and AP$_{50:95}$ values were obtained by re-evaluating the released models using the original evaluation protocol~\cite{silva2024recurrentyolov8basedframeworkeventbased}, reproducing the published AP$_{50:95}$ results.}

\compacttable
\begin{tabular}{llccccc}
\toprule
Dataset & Model & Recurrent. & Temporal context & Params & AP$_{50}$ & AP$_{50:95}$ \\
\midrule

\multirow{5}{*}{PEDRo}
& EventCenterNet (Ours) & No & 40 ms & 25M & \textbf{0.918} & 0.636 \\
& YOLOv8x~\cite{silva2024recurrentyolov8basedframeworkeventbased} & No & 40 ms & 58M & 0.895 & 0.586 \\
& ReYOLOv8n~\cite{silva2024recurrentyolov8basedframeworkeventbased} & Yes & 200 (40$\times$5) ms & 4.7M & 0.878 & 0.639 \\
& ReYOLOv8s~\cite{silva2024recurrentyolov8basedframeworkeventbased} & Yes & 200 (40$\times$5) ms & 8.3M & 0.905 & 0.655 \\
& ReYOLOv8m~\cite{silva2024recurrentyolov8basedframeworkeventbased} & Yes & 200 (40$\times$5) ms & 18.1M & 0.908 & \textbf{0.692} \\
\midrule

\multirow{4}{*}{Gen1}
& EventCenterNet (Ours) & No & 150 ms & 25M & 0.691 & 0.406 \\
& ReYOLOv8n~\cite{silva2024recurrentyolov8basedframeworkeventbased} & Yes & 550 (50$\times$11) ms & 4.7M & 0.745 & 0.464 \\
& ReYOLOv8s~\cite{silva2024recurrentyolov8basedframeworkeventbased} & Yes & 550 (50$\times$11) ms & 8.3M & 0.769 & 0.483 \\
& ReYOLOv8m~\cite{silva2024recurrentyolov8basedframeworkeventbased} & Yes & 550 (50$\times$11) ms & 18.1M & \textbf{0.780} & \textbf{0.495} \\
\bottomrule
\end{tabular}

\label{tab:pedroandgen1_comparison}
\end{table}

Table~\ref{tab:pedroandgen1_comparison} compares the proposed feed-forward EventCenterNet using the best spline-based representation, obtained from experiments in section ~\ref{seq:approx} and~\ref{seq:TemporalPlacement}, against recurrent and non-recurrent YOLO-based detectors.

AP$_{50}$ and AP$_{50:95}$ values for ReYOLO were obtained by re-evaluating the released models using the same evaluation framework as in the original paper~\cite{silva2024recurrentyolov8basedframeworkeventbased}. The reproduced AP$_{50:95}$ values matched the published results.

On PEDRo, EventCenterNet achieves the highest AP${50}$ among all compared methods and substantially outperforms the non-recurrent YOLOv8x baseline in AP$_{50:95}$. Although recurrent models achieve the highest AP$_{50:95}$, the proposed feed-forward detector remains competitive with the smallest ReYOLO model despite using no recurrent state and a shorter temporal window.

On Gen1, recurrent ReYOLO models continue to achieve the highest performance by exploiting substantially longer temporal context through hidden-state propagation. However, the improvements obtained solely through representation design suggest that a substantial portion of temporal information can be encoded directly within the event representation, while long-horizon temporal aggregation remains beneficial for challenging urban driving scenes.

\subsection{Kalman-Based Temporal Filtering}

\begin{table}[t]
\centering
\caption{Kalman-based temporal filtering on PEDRo.}
\compacttable
\begin{tabular}{lccc}
\toprule
Method & AP$_{50:95}$ & AP$_{50}$ & AP$_{75}$ \\
\midrule
Spline + local conf.      & 0.636 & \textbf{0.918} & 0.752 \\
+ Kalman, center + size   & 0.635 & 0.911 & 0.752 \\
+ Kalman, size only       & \textbf{0.639} & \textbf{0.918} & \textbf{0.757} \\
\bottomrule
\end{tabular}
\label{tab:temporal_ablation}
\end{table}

We evaluated lightweight Kalman-based temporal filtering on top of the
best spline-based baseline. Since PEDRo is recorded using a moving
camera, smoothing bounding box centers introduced localization lag due to
unmodeled ego-motion. In contrast, restricting the filtering to bounding box
width and height improved localization performance.
The results in Table~\ref{tab:temporal_ablation} show that smoothing bounding box centers does not improve localization, likely due to ego-motion-induced lag from the moving camera. Filtering only box width and height improves AP and AP$_{75}$ while preserving detector-predicted centers. This suggests that the representation already captures substantial short-term temporal continuity, while lightweight size stabilization can still provide modest localization gains.

\section{Discussion}

The experiments show that representation design substantially influences event-based object detection. Using the same fully feed-forward detector isolates this effect, with the proposed multi-timescale representations consistently outperforming compact CSTR on both PEDRo and Gen1.

The experiments further show that temporal function design is an important factor in event-based object detection. Compared with exponential decay, logarithmic B-spline encoding shows promise for localization and robustness, suggesting that the shape and placement of the temporal functions influence the temporal information retained by the representation. Parameter studies further show that PEDRo benefits from a broader distribution of temporal support within the event window, whereas Gen1 benefits from higher temporal resolution near recent events, indicating that temporal representations should be adapted to the target application rather than relying on a fixed temporal distribution.

Local confidence support provides little benefit for exponential decay but consistently improves logarithmic spline encoding on PEDRo, suggesting that temporal encoding and spatial support should be designed jointly.

The comparison with recurrent detectors provides additional insight into representation-level temporal memory. On PEDRo, the proposed feed-forward detector achieves the highest AP$_{50}$ among the compared recurrent and non-recurrent YOLO-based detectors without recurrent state. On Gen1, optimizing temporal placement and context provides clear gains, although the remaining gap to recurrent models indicates a benefit from longer temporal aggregation.

The qualitative examples also reveal several detections counted as false positives that correspond to unannotated vehicles, suggesting that missing annotations slightly underestimate the reported performance.

Kalman filtering provides only modest gains, suggesting that the proposed representation already captures most short-term temporal continuity.

Finally, the polynomial approximation retains much of the analytical spline performance while enabling recursive exponential states, with accurate approximation depending on the choice of exponential functions.
The choice and coverage of temporal functions across scenarios, their applicability to other event-based perception tasks, and efficient recursive implementations for event-driven and neuromorphic processing remain important directions for future research.


\section{Conclusion}
This work investigates how efficient multi-timescale event representations can improve event-based feed-forward object detection by encoding temporal information directly within the representation instead of relying on recurrent hidden-state propagation.

We isolated the effect of representation design and introduced a confidence-normalized continuous multi-timescale representation based on logarithmic B-spline temporal encoding, together with local confidence support and a recursive exponential-polynomial approximation.

Across both PEDRo and Gen1, the proposed representations consistently outperformed the compact CSTR baseline. On PEDRo, the spline representation improved AP$_{50:95}$ from 0.566 to 0.630 (+6.4) and increased small-object performance AP$_s$ from 0.010 to 0.062 (+5.2). 
By finetuning temporal function placement and support on Gen1, we also found that temporal representations are application dependent, improving AP$_{50:95}$ from 0.365 to 0.406 (+4.1).


As shown in Table~\ref{tab:pedroandgen1_comparison} our EventCenterNet model achieved the highest AP$_{50}$ among all compared recurrent and non-recurrent YOLO-based detectors. The achieved AP$_{50:95}$ was also competitive with recurrent ReYOLO models despite requiring no recurrent state. On Gen1, the remaining performance gap indicates that long-horizon temporal aggregation still benefits urban driving scenes. 

Finally, the polynomial approximation from recursive exponentials replaced the analytical spline representation with a compact recursive formulation while causing only minor changes in detection performance. Matching the exponential functions to the temporal support of the spline representation proved important for accurate approximation and efficient event-by-event updates.

Overall, the results demonstrate that a substantial portion of the temporal information provided by recurrent temporal modeling can be encoded directly within the event representation, providing a promising foundation for efficient feed-forward, event-driven, and future neuromorphic perception systems.

\noindent\textbf{Acknowledgments:}
This work was supported by the Wallenberg AI, Autonomous Systems and Software Program (WASP) funded by the Knut and Alice Wallenberg Foundation and Saab AB. Computational resources were provided by the National Academic Infrastructure for Supercomputing in Sweden (NAISS), funded by the Swedish Research Council.

 \clearpage

\bibliographystyle{splncs04}
\bibliography{main}

\end{document}